# 3D point cloud for objects and scenes classification, recognition, segmentation, and reconstruction: A review

**Omar Elharrouss** [1], **Kawther Hassine**[2], **Ayman Zayyan**[3], **Zakariyae Chatri**[4], **Somaya Al-Maadeed**[5], **Khalid Abualsaud**[6]

**ABSTRACT**: Three-dimensional (3D) point cloud analysis has become one of the attractive subjects in realistic imaging and machine visions due to its simplicity, flexibility and powerful capacity of visualization. Actually, the representation of scenes and buildings using 3D shapes and formats leveraged many applications among which automatic driving, scenes and objects reconstruction, etc. Nevertheless, working with this emerging type of data has been a challenging task for objects representation, scenes recognition, segmentation, and reconstruction. In this regard, a significant effort has recently been devoted to developing novel strategies, using different techniques such as deep learning models. To that end, we present in this paper a comprehensive review of existing tasks on 3D point cloud: a well-defined taxonomy of existing techniques is performed based on the nature of the adopted algorithms, application scenarios, and main objectives. Various tasks performed on 3D point could data are investigated, including objects and scenes detection, recognition, segmentation, and reconstruction. In addition, we introduce a list of used datasets, discuss respective evaluation metrics, and compare the performance of existing solutions to better inform the state-of-the-art and identify their limitations and strengths. Lastly, we elaborate on current challenges facing the subject of technology and future trends attracting considerable interest, which could be a starting point for upcoming research studies.

**KEYWORDS**: Point cloud. 3D point cloud. 3D object recognition. 3D object segmentation. 3D object and scene reconstruction.

---

[1]Department of Computer Science and Engineering, Qatar University, Doha, Qatar
[2]Higher School of Communications of Tunis/ University of Carthage, Tunis - Tunisia
[3]Department of Computer Science and Engineering, Qatar University, Doha, Qatar
[4]Department of Informatics Faculty of Sciences Dhar-Mahraz University of Sidi Mohamed Ben Abdellah, Fez, Morocco
[5]Department of Computer Science and Engineering, Qatar University, Doha, Qatar
[6]Department of Computer Science and Engineering, Qatar University, Doha, Qatar

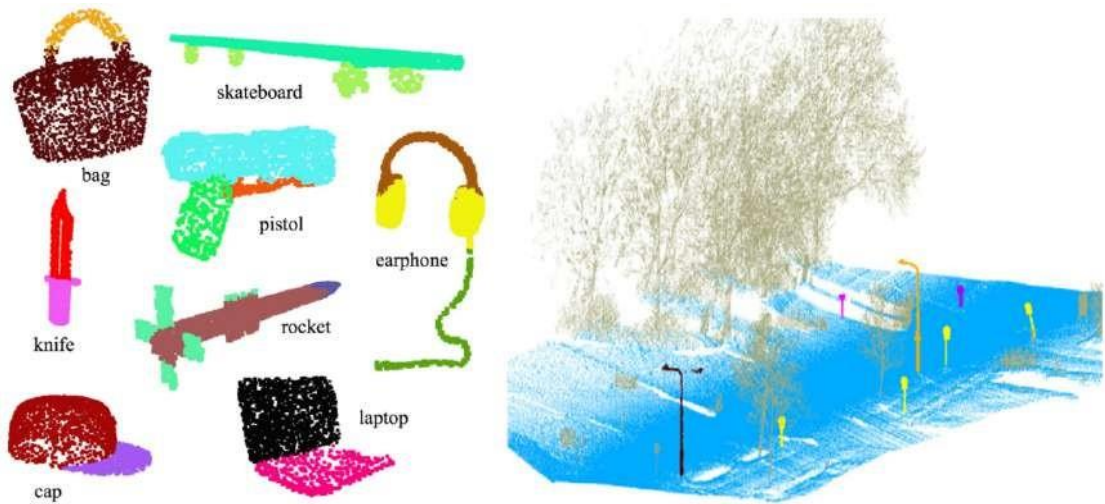

Figure 1: Object and scene with 3D point cloud representation

# 1 Introduction

The development of acquisition methods and sensing technologies contributed considerably to capturing objects and monitoring scenes using Three-Dimensional (3D) representation [1]. As a result, these tasks have gotten much easier and more cost-effective: the use of 3D images provides efficient information about the captured objects and scenes, which supports many applications such as object reconstruction, automatic driving, scene scanning, etc.

The most basic shape of a 3D representation is a Point cloud: it is formed by a group of single points drawn in a 3D area [2] to represent objects. To each point are associated several measurements, including coordinates along the X, Y, and Z axes as well as more specific data such as the color value, recorded in Red Green Blue (RGB) format [3], and the lightness measure, which determines the point's brightness.

The point cloud is captured by laser scanners or LiDAR sensors. These latter use different technologies when it comes to generating the 3D shape of objects or scenes: LiDAR sensors use various beams to capture the density of points cloud, while laser scanners use light speed to generate the 3D points cluster. The efficiency of each sensor lead depends on the number of dots being generated. This number is also a determining factor when it comes to the effectiveness of the methods working on the 3D point cloud. Figure 1 represents some examples of objects and scenes depicted in 3D point clouds.

For the human eye, shapes and objects represented in 3D point clouds are easily detected and recognized even if the density of the gathered points is sparse. For example, the swarm of dots that represents a tree can be recognized easily by the human eye, yet, a machine can consider this set of dots as noise which makes it hard to learn for the recognition algorithm. That being said, close attention was paid to finding effective methods to handle 3D point clouds growing up in a wide range of research fields [4, 5]. Fortunately, with deep learning techniques and large-scale datasets, models became more capable of learning from various features and obtaining the desired outputs; whether it is about segmentation reconstruction or recognition of 3D shapes and scenes.

Unlike triangular meshes, the point cloud does not require storing or maintaining polygonal network connectivity [6] or topological consistency [7]. In addition, point cloud technology brought significant improvements in performance and reduced overhead costs. Therefore, this research area has drawn growing interest for a wide variety of applications [8]. Several sensing technologies have been used to create 3D models from 3D Point Clouds. This involves sensors like laser scanners, LiDAR, as well as Kinect, equipped with depth sensors to provide the distance and the depth of the object.

Even though the quick development of these sensing technologies makes the extraction of point clouds much easier [9], the analysis of point clouds is still one of the key aspirations [10]. In fact, point clouds usually suffer from noise pollution and the presence of outliers. This is usually due to limitations and noise inherent in sensing devices [11] and the nature of lightning, surface reflection, or artifacts in the respective areas [12]. As a way to redress this issue, filtering was implemented on key point clouds in order to achieve accurate results suitable for further processing. Yet, most of these approaches targeted meshes and only a few studies dealt directly with point clouds [13].

3D point cloud processing, on the other hand, attracted more attention and became a subject of interest and analysis [14] worldwide. Different tasks were designed to process 3D point cloud data like, 3D object detection, recognition and segmentation as well as object and scene reconstruction [15].

In this regard, several studies have been performed, as for instance, Dong et al. in [16] proposed a new 3D object recognition method based on sequential point cloud coding. This method first transforms a point cloud into an ordered multi-channel 2D array suitable for efficient 2D convolutional operation. While authors in [17] developed a localization method that integrates a supervised object recognition method to predict probabilistic distributions on classes of objects for individual sensor measurements, into the Bayesian

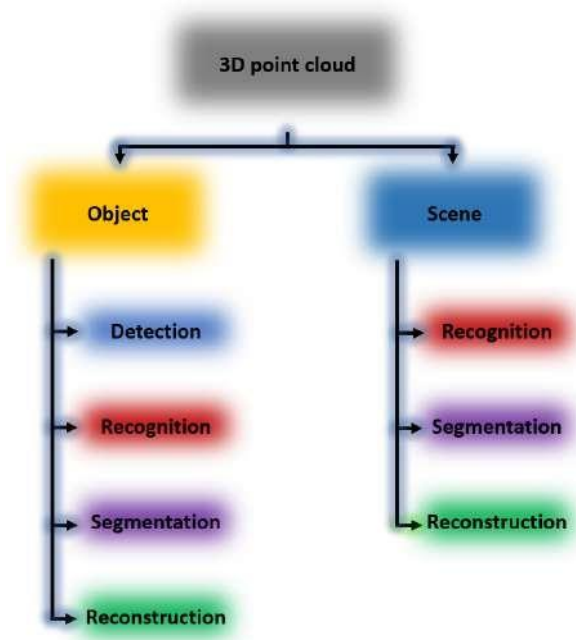

Figure 2: Computer vision tasks on 3D point cloud data

network for localization.

At this stage, reviewing different tasks and techniques pertaining to the 3D point cloud should serve as a good starting point for researchers. Actually, it helps initiate basic concepts and understanding of leading ideas of the subject matter. Even though this paper is not an in-depth investigation of 3D point Clouds field of study, it does perform an overview of all existing tasks and approaches for 3D point cloud analysis. Within, we conduct an experiment to compare the performance of several methods on current datasets. Hence, the contributions of this work could be outlined as follows:

- A thorough taxonomy of the existing works is conducted with reference to various aspects, such as the methodology deployed to design the proposed models, types of image data that could be processed by the subject technology, and their application scenarios.

- Public datasets deployed to validate the proposed models on 3D point cloud data are discussed and compared with regard to different parameters.

- Several comparisons of the most significant works identified in the state-of-the-art have been conducted to demonstrate their performance across multiple datasets and metrics.

・Current challenges and issues that remained unresolved have been targeted. Insights about future directions and trends have been discussed along with their potential impact on research and development in the near and distant future.

The remaining of this article is organized as follows: first we provide an overview of the filtering techniques applied to 3D point clouds. Then we describe the experiments performed on this kind of data before we detailed and discussed the respective results. Finally, we conclude the paper and we identify potential challenges and future perspectives.

# 2  Related Works

The impact of point clouds on many computer vision applications pushed researchers toward finding the best techniques to classify, recognize or segment the objects pictured using point clouds. In this context, many approaches have been proposed, and from which we described and analyzed the ones we presumed more relevant. These approaches are split into categories in accordance with the tasks performed on point clouds: this could be object classification and recognition, object scene segmentation, and object scene reconstruction. A summarization of the main characteristics of each method and for each task is detailed in Tables 1 and 2.

## 2.1  Objects detection and recognition

Recently, computer vision and pattern recognition communities have been focalizing on object detection. It has always been considered an essential task in computer and robot vision systems [18, 19, 20, 21]. Despite the progress achieved in this area, a lot of improvements would have to be made to achieve human-level performance. In 3D representation of data, the detection and recognition of objects represent a different, more complicated task compared to the work performed on 2D data [22, 23].

Recognizing 3D objects could be more problematic if the 3D point cloud data includes missing parts and-or noisy data. To reach accurate recognition, researchers were compelled to find methods to recover or estimate these parts of obscured data. However, the particularity of 3D point cloud representation, compared to RGB images and the usual types of data, makes it hard to operate without proper data preparation. The 3D point cloud data should undergo a series of pre-processing steps such as data denoising and cleansing before initiating the recognition process. This does not change the fact that working with 3D point clouds, instead of conventional data, has driven computer vision applications towards new strategies of data analysis and optimized consumption of memory and storage.

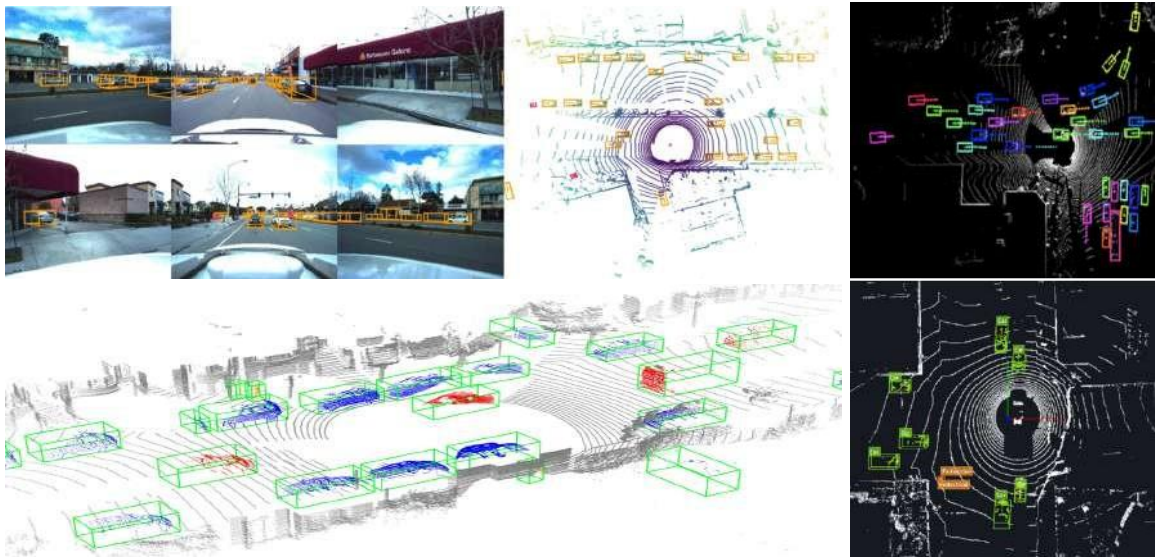

Figure 3: Car detection and recognition from 3d point cloud data

Basically, capturing 3D point cloud requires accurate sensors like laser, LiDAR, and range finder to generate precise measurements. Yet, in many cases, low-quality sensors are the ones being used. Here is where things get complicated and the need to achieve 3D object recognition despite low-quality representation arises. To handle the imperfect acquisition, authors in [28] suggested the use of descriptors (usually exploited in other computer vision domains) to process 3D point clouds. This method establishes an object recognition process based on Covariance descriptor, which proved to be effective in many similar subject studies.

In the same low-quality context, authors in [30] proposed a new object descriptor named Global Orthographic Object Descriptor (GOOD) where robots recognize objects on a real-time basis. This approach aimed real-time recognition, easy descriptive use, and low computational cost. The evaluation was performed on RGB-D Object dataset. On the other hand, authors in [31] built up an algorithm named iterative Closest Labeled Point (iCLAP) based on kinesthetic and tactile information. This solution combined two sensing modalities to create a composite perception of objects and yielded a significant improvement in recognition rate despite low-quality conditions.

Table 1: Object detection recognition and segmentation methods and their characteristics.

| Task | Method | Technique | Dataset |
|---|---|---|---|
| **Object detection and recognition** | [24] | STEM, CaRFs, VGG | Washington RGB-D, Cornell grasping |
| | [25] | PPF, LRF | UWA, Queen's, bin picking |
| | [26] | Appearance-Based Object Recognition (ABOR) | KITTI |
| | [27] | Cross-modal, SHOT and ESF descriptors | self-collected |
| | [28] | Covariance descriptor | RGB-D |
| | [29] | LonchaNet, CNN, GoogleNet | ModelNet-10 |
| | [30] | Global Orthographic Object Descriptor | Washington RGB-D |
| | [31] | iCLAP, tactile features | – |
| | [32] | 3D FCN architecture | KITTI |
| | [33] | RNN, MLP | ModelNet40 |
| | [34] | 3D CNN | ModelNet40, MNIST |
| | [12] | Statistical algorithm | self-collected |
| | [35] | 3DSmoothNet, SDV, LFR | 3DMatch, ETH |
| | [36] | voxels-based shape descriptor | self-collated MLS point cloud |
| | [37] | 3D cluster descriptor | ImageNet |
| | [38] | PointNet, NetVLAD VGG/AlexNet | Oxford RobotCar |
| | [39] | ShapeContextNet (SCN) | ModelNet40, MNIST |
| | [40] | PVNet, Embedding Attention Fusion | ModelNet40 |
| | [41] | VoxelNet, RPN | KITTI |
| | [42] | SO-Net | ModelNet10, ModelNet40, MNIST |
| | [43] | Saliency map, Point cloud conversion, Hough voting | RGB-D converted to 3d point cloud |
| | [44] | 3DCNN, MLP | S3DIS |
| | [45] | Adopted 3D ZSL | ModelNet-10-40, Mcgill, SHREC2015 |
| | [46] | TZSL, GZSL, 3D CNN, ResNet-101 | ModelNet-10-40, McGill, SHREC2015 |
| | [47] | 3DCNN, ResNet | ModelNet10, ModelNet40 |
| | [48] | GAN | KITTI |
| | [49] | RNN, K-D tree searching method | Bin-picking, UWA |
| | [50] | RS-CNN, MLP | ModelNet40 |
| | [51] | MeteorNet | MSRAction3D |
| | [52] | 2D CNN, JointNet | HDRObject9, SMDObject6, SUObjects |
| | [53] | InterpCNNs, MLP | ModelNet40 |
| | [54] | Attentional PointNet, | KITTI |
| | [55] | SFCNN | ModelNet40 |
| | [56] | CNN, YOLO | KITTI |
| | [57] | PointNet++, PGM, PointsPool | KITTI |
| | [58] | DGCNN | ModelNet40 |
| | [59] | MVF, CNN | KITTI |
| | [60] | Point pair feature | UWA, Real scene |
| | [61] | Octree, DeepCNN, VoxRec | Shapenetcore Part |
| | [62] | two-stage Weakly Supervised | KITTI |
| | [63] | VoxNet, 3DCNN | RGB-D, LiDAR |
| | [64] | 3D to 2D projection, Inception-v3 | KITTI |
| | [65] | MVG, RPN | KITTI |
| | [66] | RPN, segmentation network, estimation network | KITTI |
| **Object segmentation** | [26] | CNN, Contextual Labeling Refinement (CLR) | KITTI |
| | [67] | Voxel-based 3D-CNN | LiDAR point cloud Ottawa |
| | [33] | 3D CNN, MLP | ShapeNet part |
| | [34] | 3D CNN | ModelNet40 |
| | [37] | 2D CNN labeling | ImageNet |
| | [39] | Attentional ShapeContextNet (A-SCN) | ShapeNet part |
| | [42] | SO-Net | ShapeNet part |
| | [50] | RS-CNN, MLP | ShapeNet part |
| | [51] | MeteorNet | KITTI, Synthia |
| | [53] | InterpCNNs, MLPs | ShapeNet Part |
| | [55] | SFCNN | ShapeNet Part |
| | [58] | DGCNN | ShapeNet Part |
| | [61] | Octree, DeepCNN, VoxRec | Shapenetcore Part |

For the same purpose, and using hierarchical cascading for low-quality, authors in [24] introduced an object recognition and detection solution from RGB-D images. This method consists of analyzing comprehension probabilities and object class probabilities by extracting features using VGG architecture. Also, to recognize objects using a unified representation based on point clouds, authors in [27] proposed a cross-modal Visio-Tactile object recognition method using tactile data retrieved from a tactile skin reader. SHOT and ESF descriptors were used for the recognition process.

Other studies tackled differently noisy and partial data in 3D point clouds. For instance, authors in [25] implemented an object detection/recognition method based on Local Reference Frames (LRFs) and Point Pair Feature (PPF). The evaluation was performed on three datasets including UWA, Queen's, and Bin-Picking. In [29] authors designed a CNN-based architecture named LonchaNet for real-time recognition represented in 3D point cloud. The proposed model exploited GoogleNet network for features extraction and was tested on ModelNet-10 dataset. On the same dataset, authors in [63] developed a 3DCNN model referred to as VoxNet, for 3D object recognition. On KITTI dataset, authors in [32] adopted a 3D vehicle detection using 3D boundary boxes by extending 2D FCN architecture to 3D FCN. Another low-quality case is camera self-occlusions caused by hand obstruction. To overcome object obstruction, authors in [12] developed a hand gesture recognition method using data 3D point could data. This method is based on statistical and probabilistic theories.

For classification purposes, authors in [33] designed an RNN architecture named PointNet using Multi-layer perception (MLP). Their purpose was to overcome small perturbations of input points as well as errors through point insertion or deletion. To improve the performance of PointNet architecture, authors in [34] used as metric space distances to enable learning from local features as well as increasing contextual scales by learning from multiple scales. Authors in [36] used mobile laser scanning point cloud data to design 3D multi-scale shape descriptor for object recognition named SigVox. This method used a shape description exploiting voxels to identify different types of lamp poles and traffic signs in streets.

In [37] authors' main objective was to create classification solutions for 3D object recognition from point cloud data. They implemented 2D Convolutional Neural Network (CNN) for labeling, 3D clustering descriptor for object segmentation, and a deep learning-based model for recognition. Another 3D object recognition method was proposed in [39] which involves a deep learning-based technique named Attention ShapeContextNet (SCN). The architecture concatenates a list of ShapeContextNet blocks followed by Multilayer Perceptron (MLP) layers for the final classification. The proposed method has been evaluated on ModelNet40 classification dataset. With the same leading idea, authors in [40] used a deep-learning-based method for 3D shape recognition over point cloud data and multi-view 3D shape representation of the object. This combination was executed through a unified network while an Embedding Attention Fusion (EAF) block was used over the features on point cloud and multi-view 3D shape; features extraction was carried out using AlexNet backbone.

Detecting 3D objects from LiDAR point cloud data was the focus in [41]. In this work, authors introduced a generic 3D detection network named VoxelNet which consists of extracting features before predicting boundary boxes. Region proposal network (RPN) was adapted to interface with the 3D point cloud while generating boundary boxes and probability maps. Authors in [42] proposed a 3D point cloud classification method named SO-Net. SO-Net architecture is a multi-task network that extracts global features to be used in the classification network. ModelNet10, ModelNet40, and MNIST datasets were used for assessment purposes. In [35] proposed a compact learned local feature descriptors for 3D point cloud matching. They introduced the smoothed density value (SDV) voxelization as a novel input data representation in order to reduce the sparsity of the input voxel grid, mitigate the boundary effects and smooth out miss-alignments due to local reference frame (LRF) estimation errors.

The unstructured distribution of LiDAR point clouds has also been an issue when it comes to pattern learning [68]. Henceforth, authors in [69] proposed a variant of Broad Learning System (BLS) that implements a Unified Space Auto-Encoder (USAE). Referred to as USAE-BLS, the result came out as a lightweight model for 3D object recognition. In [70] authors sought object classification from 3d point cloud representation. While combining point clouds and voxelized data, they used the Euclidean distance between the point and the object's center to extract connections between points within each voxel. Next, a deep learning network performed the task of object classification. A different deep learning-based method named Drop Channel Graph Neural Network (DC-GNN) was designed in [71] to extract meaningful features and depict an effective representation of objects before classification. This solution relied on k-NN-based drop channel and Multi-Layer Perceptron (MLP) Networks. In [72], another MLP-based method implemented spatial-shift MLP as the backbone to determine the relationship between patches from different views. A different approach, known as view set pooling transformer (OVPT), [73], was designed to build a comprehensive image of the same object from different perspectives using view information entropy computation.

In order to recognize objects from RGB-D images, a segmentation process was performed on RGB and in-depth images [43]. This step was followed by a conversion and a generation to 3D point cloud representation. On resulting 3D point cloud data, features extraction followed by Hough voting operation are performed to recognize existing objects. Using 3D laser-scanned point clouds, authors in [44] established a 3DCNN-based deep learning method (MLP used in the last layer) to recognize building elements and perform object reconstruction. To evaluate the model, authors used the S3DIS dataset. Exploiting Generative Adversarial Network (GAN) was also one of the ways to handle 3D point cloud object recognition: authors in [48] proposed a method called Point Cloud Up-sampling Adversarial Network,**PU-GAN** which consists in identifying an object from a whole scene captured using LiDAR sensors. Considering that the scene may comprise other objects like buildings, trees, etc. PU-GAN is designed to fill the detected object by up-sampling the point sets. In [49], instead of transforming point clouds to voxel grids, authors introduced a technique of extracting 3D key point features directly from point clouds. These features are used to recognize this object by exploiting and preserving its geometric information. In a different work, [50], a Relation-Shape Convolutional Neural Network (RS-CNN)-based solution was devised for object classification, segmentation, and reconstruction. Authors in [53] proposed a CNN-based deep learning model named Interpolated Convolution Network (InterpConv). InterpConv was implemented for object recognition and segmentation purposes. While authors in [54] relied on PointNet architecture to detect and recognize 3D objects. In [47] authors used ResNet backbone for features extraction on multi-view of 3D presentation of an object, afterward they implemented a deep-learning-based approach named View-based Weight Network (VWN) for 3D object recognition. Projections of different views of a given object were used for its identification. Besides, to classify, segment and estimate the scene flow from 3D point clouds, authors in [51] introduced a method named MeteorNet. Unlike previous works which used normal 3D point cloud vectors, MeteorNet's purpose was to classify objects from 3D point cloud sequences using KITTI dataset for performance assessment.

Table 2: Summarization of scenes recognition and objects reconstruction methods

| Task | Method | Technique | Dataset |
|---|---|---|---|
| Scenes and places recognition | [74] | PointNetVLAD, LPD-Net, RN-VLAD, GCA | Oxford RobotCar |
| | [38] | PointNet, NetVLAD VGG/AlexNet | Oxford RobotCar |
| | [66] | RPN, segmentation network, estimation network | KITTI |
| Scenes and obejcts reconstruction | [75] | Mesh deformation network and Graph Unpooling Network | ShapeNet |
| | [76] | point set generation network (PSGNet) | ShapeNet |
| | [77] | Point Cloud Reconstruction Network, Silhouette Completion Network | ShapNet |
| | [78] | CNN | ShapNet |
| | [79] | Encoder-decoder | ShapNetCore, Pix3D |
| | [80] | RfD-Net, 3D detector, spatial transformer, shape generation | ScanNet [81],Scan2CAD [82] |
| | [83] | Building Information Modeling (BIM) | 2D-3D-S, S3DIS |
| | [84] | unsupervised Building Information Modeling (BIM) | 2D-3D-S, S3DIS |
| | [85] | Vectorized model, Multistep 2D optimization | MC and Office |
| | [86] | CNN, building element segmentation and reconstruction | S3DIS,2D-3D-S |

Despite the wide scope of solutions for object recognition, a concern about identifying unseen classes of 3D point cloud data leads to using some of the techniques initially applied to 2D data: one of which is Zero-Shot Learning (ZSL). For instance, authors in [45] adopted a 2D ZSL method to 3D point cloud data by changing the setting parameters of the original ZSL approach. In that same regard, authors in [46] combined Transductive Zero Shot Learning (TZSL) and Generalized Zero-Shot Learning (GZSL) techniques for 3D point cloud classification.

Even though most of the existing methods operated directly on 3D point cloud representation for object recognition, other approaches chose to convert, first, 3D data into 2D data. That is, to benefit from the existing object recognition methods with deep learning techniques. For instance, authors in [52] developed a solution for object recognition using 2D representation while converting 3D shape of point clouds to 2D multi-view representation cloud data. Then, work on the new data with a 2D CNN-based method to recognize objects. A fusion module is implemented to collect different multi-view features for classification parts. The proposed method was tested using three datasets namely HDRObject9, SMDObject6, and SUObjects. Another paper, [87], investigated an alternative approach, which is based on the alignment of convex hulls of segments detected in a depth image with convex hulls of target 3D object models. The input of the algorithm is a triangular mesh obtained from 3D point cloud and the output is a set $C$ of convex surfaces $C_i$. Also, by converting 3D point cloud data to a 2D representation, authors in [64] used the projection technique for mapping 3D data into 2D image. Then

Inception-v3 was used for data classification. Object recognition is performed using the fusion of image-based detection and point cloud data. Also, using an adaptive spherical projection of 3D point cloud with a fractal structure, authors in [55] implemented a convolutional neural network for 3D object recognition. Using 2D representation, the spherical projection enables the deep model to learn from various features.

From a sensing perspective, it is commonly known that methods based on LiDAR information are good to detect obstacles, yet limited for other types of objects. For better outcomes, combining the vision and point cloud (LiDAR) information is used in [56] by exploiting a fusion method for vehicle detection. In this solution, obstacles are detected using point cloud data, then mapped to the image. The YOLO technique is used for the detection and localization of vehicles. While the boundary boxes are mapped to the same image. In [57], authors defined a two-stage 3D object detection framework, named sparse-to-dense 3D Object Detector (STD). The proposed network goes through two stages: PointNet++, as backbone, detects the regions where the objects are, then PointsPool uses the first stage's features to detect objects with a compact representation.

In another research work namely, [59], authors designed a multi-view Fusion (MVF) method by introducing a dynamic voxelization algorithm with the purpose of overcoming the problems affecting existing Voxel-based methods. The proposed voxelization method consists of two branches in the MVF network including the Bird-eye view and perspective view used for features extraction. From each branch, a list of convolution and pooling layers is fused to obtain the final classification. Using a new descriptor based on point pair features of 3D point cloud objects, authors in [60] estimated and recognized the poses of an object. Authors in [61] used many features such as surface normal and surface curvature and 3D point cloud for classification and segmentation processes. The authors proposed a multitask-deep-learning-based method by extracting the histogram and voxelized features of an object to be used as input for the proposed CNN network. Authors in [62] operated on LiDAR 3D point cloud data and implemented a two-stage weakly supervised system for 3D object detection. This model is trained on a small dataset with few annotated scenes (500 annotated scenes). The first stage operates on cylindrical object generation learning while the second stage implements cuboids confidence score selection. Authors in [65] implemented Multiple Views Generator (MVG) methods that observe the scene from four views, and a region proposal network (RPN) for object detection. While In [88]

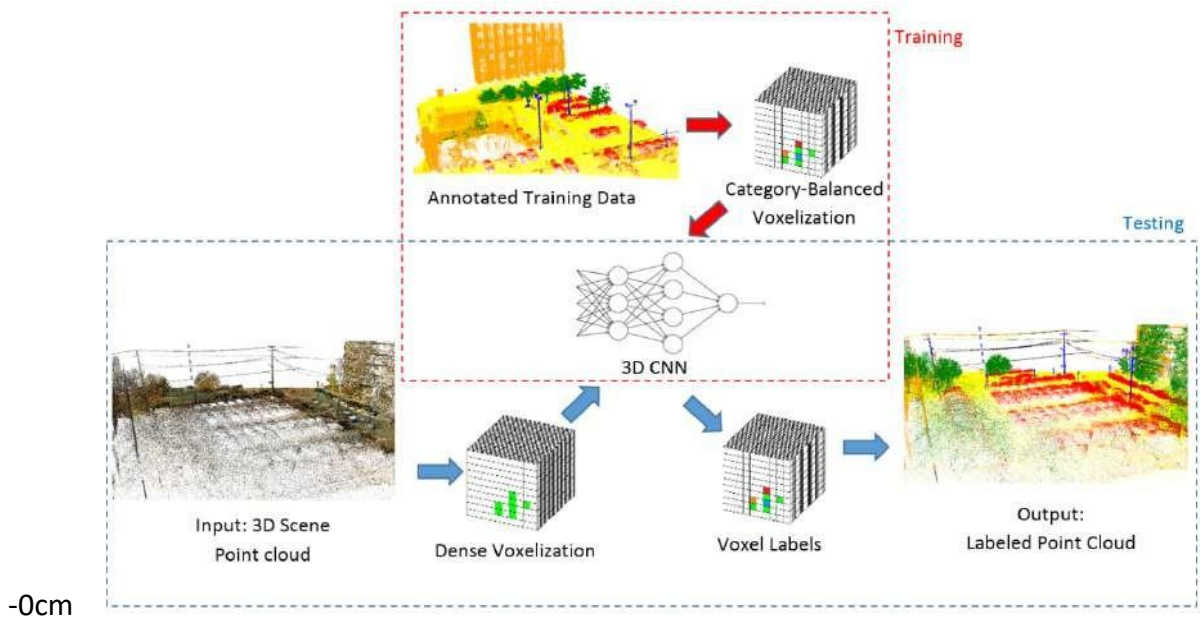

-0cm

Figure 4: 3D point cloud segmentation proposed in [67]

authors proposed a 3D local feature descriptor named Histograms of Point Pair Features (HoPPF) for 3D object recognition. The HoPPF method implies robust representation and efficient computation.

In [89] authors proposed a dynamic ball query (DBQ) network for subset selection of input points and then assigned the feature transform to each selected point. Yet, in many cases, objects are partially covered and the information gathered is barely sufficient for accurate detection. Accordingly, authors in [90, 91] presented a LiDAR-based Behind the Curtain Detector (BtcDet) model to complete the shape of objects before detection. By learning the shape priors, this solution estimates the complete shape of partially obscured (curtained) objects. Authors in [92] introduced a Boundary-Aware 3D Object Detection BADet network. In the form of a local neighborhood graph, this method efficiently displays local boundary correlations of an object. For more informative RoI-wise representations, a lightweight Region Feature Aggregation Module was devised to exploit voxel-wise, pixel-wise, and point-wise features with expanding receptive fields. In [93, 94] authors described a solution for 3D object detection and recognition based on two different inputs of the same object: the RGB image and the 3D point cloud. On the other hand, authors in [93] implemented a network for 3D object detection using encoder-decoder architecture, a refinement module, and a Cascade Bi-directional Fusion.

## 2.2  Scenes and Place recognition

Recognizing the observed scenes could be very helpful for navigation and localization purposes.In the case of conventional images and videos, the recognition of places and scenes could be easy but with 3D data presentation it becomes quite difficult: recognizing scenes and places from RBG images is easy using deep-learning-based techniques. But the representation of these scenes and places using 3D points makes it harder due to the potential presence of a swarm of dots as well as objects within the scene. To avoid possible confusion, many studies were conducted to differentiate places and scenes from other types of data.

Based on large-scale places datasets, authors in [38] introduced a 3D point cloud scene recognition approach named PointNetVLAD. The model encompasses three blocks, namely PointNet architecture, NetVLAD, and a fully connected network. NetVLAD architecture is used to aggregate image features extracted using VGG/AlexNet into a global descriptor vector. In [74], authors propose the Point cloud and Image Collaboration Network (PIC-Net), which It combines the features of an image with those of a point cloud and mines the complementary information they provide. Additionally, the night image is converted into a daytime image in order to improve recognition performance at night. While in [66], authors depicted a 3-step Deep Neural Network (DNN) using RGB image and LiDAR data. RPN model is put in place to generate features maps from RGB images. These features are lifted into 3D proposals to be used for segmenting object points. The estimation network is used to extract 3D objects bounding boxes.

## 2.3  Objects and scenes segmentation

Segmenting objects in a scene is convoluted when the data is captured using LIDAR sensors [95]. This data is presented with a swarm of dots and makes it hard to find a pattern to segment these objects. With the aim of labeling a scene that involves objects pictured with 3D point clouds, authors in [67] proposed an object segmentation mechanism centered on 3D convolutional neural network. This approach minimizes prior knowledge of the labeling problem and does not require a segmentation step or artisanal features. In the same context, and to segment the point cloud into regions of ground: low foreground (i.e. short objects), and high foreground (tall objects), authors in [26] built an object segmentation solution using CNN and Contextual Labeling Refinement (CLR) to complete the connected components of an object. As proposed for classification purposes, authors in [33] also presented a 3DCNN architecture for object part segmentation. The architecture is simple with 19 convolutional layers and assessed using ShapeeNet part dataset.

A new upgraded version of PointNet architecture was described in [34] and labeled PointNet++. This new version consists of Multi-scale group (MSG) and Multi-resolution group (MRG) layers and provides a comprehensive solution for part segmentation. Also, for recognition purposes, the solution in [37] was designed to segment objects for easier recognition using 3D spatial information and a CNN-based model. Meanwhile, in [39], authors adopted ShapeContextNet (SCN) for point cloud recognition and attention ShapeContextNet (A-SCN) for segmentation. The evaluation of A-SCN was performed on ShapeNet part segmentation dataset.

In [42], a 3D point cloud classification and segmentation system, SO-Net, was developed upon a multi-task network that extracts spatial distribution features before segmenting objects by a segmentation network. For evaluation purposes, authors used ShapeNet part dataset as well. Another method was devised in [50] using a deep learning architecture named Relation-Shape Convolutional Neural Network (RS-CNN). This solution was tailored to segmentation as well as classification and object reconstruction. In [51] authors presented a deep learning method named MeteorNet. Here, semantic segmentation was performed using Synthia and KITTI datasets. In [53] another deep learning model named Interpolated Convolution operation(InterpConv) was deployed to tackle point cloud feature learning and understand issues of object segmentation.

As mentioned earlier, the conversion of 3D point clouds to a 2D representation can make the processing of a deep learning model much more effective. For this reason, an adaptive spherical projection using 2D CNN model, also referred to as Spherical Fractal Convolutional Neural Networks (SFCNN), is used before segmenting objects [55]. Also, to Work with convolutional neural networks on point cloud data, an adaptation is made by authors in [58] using a spatial transformation for point cloud segmentation and classification. Authors proposed a deep learning module named EdgeConv which was replicated between the model's layers. Unlike standard methods implementing deep learning and using 3D point cloud for object recognition, authors in [61] utilized different features such as surface normal and surface curvature to classify and segment 3D objects. The authors proposed a deep-learning-based method using OCTREE encoder features for segmenting different parts of the object.

Authors of [96] proposed a 3D object segmentation method using the Dual Attention Network (DANet) to overcome the irregularity and sparsity of point clouds for 3D object classification and segmentation. DANet combines two modules: a Local Feature Extraction module (LFE) for local feature learning and a Global Feature Fusion module (GFF) for global information fusion. In [97] authors adopted a Graph Convolutional Neural Network (GCNN) to learn from contextual-point relationships between neighboring points for a structured 3D point cloud segmentation. Also using a GCNN-based network, authors in [98] attempted to locate the attentional regions using spatial information with different distances.

A Deep Feature Transformation Network (DFT-Net) was proposed in [99]. It was intended to capture local features and exploit the relationship between neighborhood points. In the same context, and in order to extract spatial and semantic consistencies between point cloud data, authors in [100] proposed a deep learning-based network that encompasses a data augmentation module and a segmentation network.

## 2.4 3D Objects, scenes and buildings reconstruction

The reconstruction of objects and scenes using different types of data including RGB-D images, Lidar, and Laser scanner data is a complex yet beneficial task for many applications such as medical imaging, virtual reality, computer imaging, etc. The reconstruction of objects, scenes, and buildings after being transferred makes the exchanges of this type of data very helpful compared to other types of data like images or videos.

Whether it is 3D laser scanning, RGB-D camera, stereo camera, or LiDAR, 3D point clouds usually represent 3D surfaces or objects in accurate manners [101]. Using 3D point clouds, a building or a surface could be pictured and reconstructed. Accordingly, multiple studies have been conducted on 3D Objects, scenes, and buildings reconstruction. For example, to generate 3D point cloud from RGB images, authors in [76] implemented a deep learning method named point set generation network (PSGNet). In the same context, yet to create 3D point objects from a single image, authors in [75] developed a CNN-based model named Pixel2Mesh with a Mesh deformation network and a Graph

Unpooling Network. In [79] an encoder-decoder network was deployed to produce RGB information in latent space and then use it to reconstruct 3D objects. Also, authors in [77] proposed a method for object reconstruction using RGB images and a silhouette map. An encoder-decoder network was dedicated to the extraction of the complete object silhouette. The completed silhouette and RGB images are introduced as input of ResNet-50 network. A reconstruction network was set up to build the point cloud representation of the object. The training and the validation were completed on ShapeNet dataset using Chamfer Distance (CD) and Earth mover's distance (EMD) metrics. Another method to reconstruct 3D object from a single image is proposed in [78]. The method named pixel2point exploited an encoder-decoder network to generate a point cloud.

The reconstruction of different components of a scene including objects, walls and buildings has become a subject of interest for many applications including interior design and robot navigation. For this reason, many researchers strived to improve reconstruction mechanisms over 3D point clouds. For example, using point cloud, authors in [80] presented a deep learning solution named RfD-Net for the detection and reconstruction of object surfaces in a scene. The detected objects were reconstructed by filling the spaces between points. With the same scope, authors in [83] introduced a method for building Walls reconstruction using Building Information Modeling (BIM). Using the same method type, authors in [84] devised an unsupervised Building Information Modeling (BIM). This method, also referred to as scan-to-BIM, consists of reconstructing the building including walls for scanned 3D point cloud data.

However, generating data using image-based 3D reconstruction or using LiDAR is prone to errors, missing parts or noisy points. Using Multi-step 2D optimization and Levels of Detail (LoDs) which performs a progressive vectorized reconstruction of models, authors in [85] reconstructed surfaces from 3D point cloud data. To recognize the scene components and reconstruct it from 3D point cloud data, authors in [86] proposed a deep learning method to detect and recognize building elements from laser-scanned data. Then the detected elements are segmented before being reconstructed.

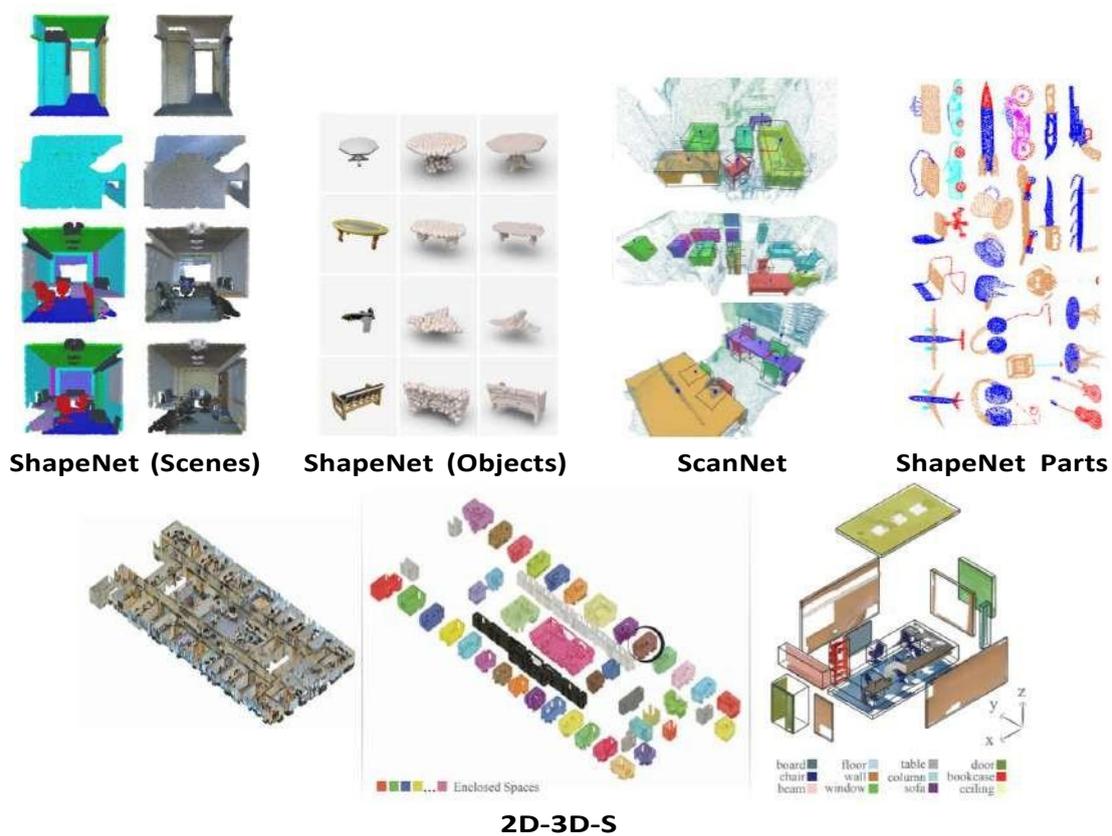

Figure 5: Sample examples from 3D point cloud datasets

# 3 Datasets

The growth of ML and DL algorithms led to the creation of large-scale datasets. These datasets have had a big impact on models outcomes and effectiveness. A dataset with thousands or billions of annotated data offers a huge amount of information models can learn from. For example, ImageNet [102] helps to evaluate visual recognition algorithms, while VGGFace [103] is designed to validate face recognition methods. For 3D point clouds, KITTI, ModelNet-10, and ModelNet-40 are the most famous data sets for object classification and recognition. For 3D object segmentation, a more recent research area, few datasets include annotation. Figure 5 portrays samples from the most famous datasets used for 3D object recognition, segmentation, and reconstruction. In the following paragraph, we will briefly discuss the individual properties of these datasets.

**KITTI** [7]**:** The images pertaining to this dataset are captured from various places of the metropolis of Karlsruhe (Germany), including highways and rural regions. It contains more than 12K images illustrating varied objects. In each image, we can find around 30

---
[7]http://www.cvlibs.net/datasets/kitti/

pedestrians and 15 vehicles.

**MSRAction3D**[8] is an action recognition dataset or 3D point representation contains in-depth images grouped in 567 map sequences with a resolution of 640x240 pixels. Images are captured using Kinect sensor.

**ModelNet:**[9] ModelNet dataset derives a set of version including ModelNet10 and ModelNet40. ModelNet10 is composed of 10 classes of different 3D point cloud objects. While ModelNet40 contains 40 classes. These two datasets are used by multiple deep-learning methods for 3D object classification and recognition.

**3D ShapeNets** [10] ShapeNet Parts is a 3D object segmentation dataset composed of more than 31K meshes of 16 classes of objects. Object shapes comprise 2 to 5 labeled parts.

**ImageNet**[11] This data set is dedicated to Large Scale Visual Recognition Challenge (ILSVRC). ImageNet is also a large-scale dataset with thousands of images and subnets. Each subnet is represented by 1000 images. The current version of the dataset contains more than 14,197,122 images wherein 1,034,908 images of the human body are annotated with bounding boxes.

**2D-3D-S:**[12] is a dataset including 2D and 3D data with segmented geometric and semantic annotations. This dataset is collected from 6 large-scale areas of more than 6000 m2 buildings with annotated 3D point cloud representations. It does, also, contains 70K RGB depth images of the same scenes.

# 4   Experimental results and discussion

In this section, we present the experimental results of the state-of-the-art works evaluated on standard 3D point cloud benchmarks, including KITTI, ShapeNet, ShapeNet parts, ModelNet10, ModelNet40, and MNIST. The accuracy and efficiency are compared between methods performing the same task. Methods performance is often measured on the validation set and test set for each dataset: if so, this distinction is taken into consideration in our comparison.

---

[8]https://www.uow.edu.au/ wanqing/#Datasets
[9]https://modelnet.cs.princeton.edu
[10]https://shapenet.org
[11]http://image-net.org/
[12]https://github.com/alexsax/2D-3D-Semantics

## 4.1 Object detection and recognition evaluation on KITTI dataset

The 3D object detection and recognition have been evaluated on the KITTI dataset. Some researchers chose to evaluate their method on validation sets only while others opted for validation and test sets like VoxelNet in [41]. Results obtained for each method are assessed using Average Precision metric (AP) on two modalities including 3D car and BEV Car. Table 3 represents respective results for validation and test sets.

On the KITTI validation set, we note that the approaches under investigation: [41], [59], [89], and [89], succeeded to detect 3D objects with close results for easy, moderate, and hard metrics in 3D car modality. However, results of [54] are 32% below the utmost one, for the first metric (Easy). For the BEV Car, only two methods investigated this modality, and the outcome showed improvements as per the methods evaluated on 3D car modality. For example, in [62] the accuracy for 3D car(easy) is 84.04% whilst it hits 90.11% for BEV Car, with a difference of 6%.

On the KITTI test set, we notice that more methods performed the evaluation for BEV Car as opposed to the validation set. For 3D car modality, EPNet++ [93] reached the highest performance for easy and moderate modalities, while it came in second place for hard modality with a precision of 76.71. For BEV car modality, BADet [92] achieved the highest precision results for easy modality with 95.23 outperforming PointRCNN+SAT-GCN [104] by 2%. The same observation applies to moderate categories. Meanwhile, results recorded by [57] are the best for hard metric.

The Average Precision metric (AP) is also considered for evaluation on Pedestrian and Cyclist modalities according to the same difficulty flags: easy, moderate, and hard. Table 4 details the results of different methods. Comparing the evaluation results on 3D Car and BEV Car, we note that the AP rates on Pedestrian and Cyclist fall behind those registered for 3D Car and BEV Car modalities. This difference demonstrates that Pedestrians and Cyclists are more challenging to process due to intrinsic characteristics and the complexity of these modalities compared to the Car modality. This makes real-time usage hard to be implemented at the present.

With reference to the same table (Table 4), we observe that the solution detailed in [57] yields better results in terms of AP values on the test set, and outperforms EPNet++

Table 3: Evaluation results of car detection on KITTI dataset

| Dataset | Test/Val | Method | 3D Car | | | BEV Car | | |
|---|---|---|---|---|---|---|---|---|
| | | | Easy | Moderate | Hard | Easy | Moderate | Hard |
| KITTI | val | Yang et al. [65] | 86.31 | 77.32 | 73.21 | 89.36 | 81.52 | 79.62 |
| | | Li et al. [32] | 84.20 | 75.3 | 68.0 | - | - | - |
| | | Meng et al. [62] | 84.04 | 75.10 | 73.29 | 90.11 | 84.02 | 76.97 |
| | | A-PointNet[54] | 58.62 | 52.28 | 47.23 | - | - | - |
| | | VoxelNet [41] | 81.97 | 65.46 | 62.85 | - | - | - |
| | | MVF [59] | 90.23 | 79.12 | 76.43 | - | - | - |
| | | DBQ-SSD [89] | 89.60 | 79.56 | 78.51 | - | - | - |
| | | BtcDet[90] | 93.15 | 86.28 | 83.86 | - | - | - |
| | test | VoxelNet [41] | 77.47 | 65.11 | 57.73 | - | - | - |
| | | Wang et al. [56] | 70.58 | 62.71 | 55.17 | - | - | - |
| | | Zhang et al. [66] | 80.25 | 69.21 | 60.15 | - | - | - |
| | | Meng et al. [62] | 80.15 | 69.61 | 63.71 | 90.11 | 84.02 | 76.9 |
| | | Yang et al. [57] | 86.61 | 77.63 | 76.06 | 89.66 | 87.76 | 86.89 |
| | | BADet [92] | 89.28 | 81.61 | 76.58 | 95.23 | 91.32 | 86.48 |
| | | Li et al. [94] | 78.68 | 61.46 | 60.09 | 88.07 | 86.26 | 69.79 |
| | | SIENet [91] | 88.22 | 81.71 | 77.22 | 92.38 | 88.65 | 86.03 |
| | | EPNet++[93] | 91.37 | 81.96 | 76.71 | - | - | - |
| | | PointRCNN+SAT-GCN [104] | 86.55 | 78.12 | 73.72 | 92.83 | 88.06 | 83.51 |

[93] by a difference under 1% for Pedestrian and 7-13% for Cyclist modality. On Val set, the method in [62] hits the best AP values for Pedestrian modality yet has not been performed on Cyclist. On the other hand, the method in [57] reached the second-best outcome for Pedestrians and SIENet [91] reach the highest one for cyclists.

## 4.2 Object shape classification On ModelNet10, ModelNet40, and MNIST datasets

To perform 3D object classification from 3D point cloud data, the referred methods involved three public datasets namely ModelNet10, ModelNet50, and 3D MNIST. Relevant results are shown in Table 5, for ModelNet10 and ModelNet40 datasets, and in Table 6 for the MNIST dataset.

The results on the ModelNet10 (displayed in Table 5) demonstrate that most methods succeeded to classify 3D objects with approximate accuracy values except for TZSL [46] Which only reached 46.9%. The best results were recorded by SO-Net [42] and VWN [47], and they outperformed TZSL by a large margin of 49%. Results achieved by Lonchanet [29], VoxNet [63], and SFCNN [55] are mostly close: the difference between remained below 2%. This proves that the aforementioned approaches were fairly efficient and accurate in terms of 3D object recognition, yet there is always scope for improvement.

Other methods chose ModelNet40 dataset for their evaluation as shown in Table 5. Respective results indicate that VWN reached the highest accuracy value of 93.8% while RS-CNN and DC-GNN achieved 93.6% with a difference of 0.2%. The same difference was recorded between RS-CNN and SO-Net [42], PVNet and SO-Net. Also, we notice that the difference between the highest and the lowest results stayed under 4% which is between VWN, VoxNet, and DC-GNN. This demonstrates the effectiveness of these solutions as well as the complexity of ModelNet40 compared to ModelNet10.

Table 4: Evaluation results for Pedestrian and Cyclist on KITTI dataset.

| Dataset | Test/Val | Method | Pedestrian | | | Cyclist | | |
|---|---|---|---|---|---|---|---|---|
| | | | Easy | Moderate | Hard | Easy | Moderate | Hard |
| KITTI | test | Zhang et al. [66] | 51.04 | 43.23 | 40.77 | 67.26 | 55.34 | 45.26 |
| | | VoxelNet [41] | 39.48 | 33.69 | 31.51 | 61.22 | 48.36 | 44.37 |
| | | Yang et al. [57] | 53.08 | 44.24 | 41.97 | 78.89 | 62.53 | 55.77 |
| | | Li et al. [94] | 39.11 | 34.42 | 27.13 | 62.09 | 43.65 | 36.80 |
| | | SIENet [91] | - | - | - | 83.00 | 67.61 | 60.09 |
| | | EPNet++ [93] | 52.79 | 44.38 | 41.29 | 76.15 | 59.71 | 53.67 |
| | val | VoxelNet [41] | 57.86 | 53.42 | 48.87 | 67.17 | 47.65 | 45.11 |
| | | Meng et al. [62] | 74.65 | 69.96 | 66.49 | - | - | - |
| | | BtcDet [90] | 69.39 | 61.19 | 55.86 | 91.45 | 74.70 | 70.08 |

For the Multi-view-based approaches that were evaluated on ModelNet10 and ModelNet40 datasets are presented in Table 5. Within, we presented obtained accuracies while taking into account the number of views in the training process. We observe that R2-MLP [72] achieved the highest performance accuracies for both datasets using 20 views. While OVPT [105] method reached the second place with an accuracy of 79.2% on ModeNet10 and 99.3% on ModNet40 using 9 views. Using 12 views ReINView [73] reached high-performance accuracy for both datasets. In terms of number of views, we can de- duce that OVPT [105] reached satisfactory results with less number of views compared to R2-MLP [72] and ReINView [73].

Table 6 presents object classification of 3D MINST dataset. The table shows that SO-Net [42] achieved better performance in terms of error rate, followed by PointNet++ [34] with 0.51 as the error rate and a difference of 0.1%. These results were produced using an input size of 512. Methods with higher error rates used an input data size of 256. This demonstrates that the size of input data can affect system effectiveness on 3D MNIST dataset.

Table 5: Object shape classification results of methods On ModNet10 and ModNet40

| Type | Method | ModeNet10 | ModNet40 |
|---|---|---|---|
| Point | TZSL [46] | 46.9 | - |
| | Lonchanet [29] | 93.3 | - |
| | VWN [47] | 95.1 | 93.8 |
| | SO-Net [42] | 95.7 | 93.4 |
| | VoxNet [63] | 93.9 | 89.7 |
| | RS-CNN [50] | - | 93.6 |
| | PointNet++ [34] | - | 91.9 |
| | Mao et al. [53] | - | 93.0 |
| | SFCNN [55] | 91.4 | 92.3 |
| | Xie et al. [39] | - | 90.0 |
| | PVNet [40] AlexNet | - | 93.2 |
| | DGCNN [58] | - | 93.5 |
| | SCN [39] | - | 90.0 |
| | USAE-BLS [69] | 92.6 | - |
| | 3DHMNN [68] | 90.9 | 83.8 |
| | Gezawa et al. [70] | 93.4 | 88.2 |
| | DC-GNN [71] | - | 93.6 |
| View | R2-MLP (20 views) [72] | 97.7 | 99.6 |
| | OVPT (9 views) [105] | 97.2 | 99.3 |
| | ReINView (12 views) [73] | 96.1 | 97.8 |

Table 6: MNIST digit classification

| Method | Input size | Error rate |
|---|---|---|
| PointNet [33] | 256*2 | 0.78 |
| PointNet++ [34] | 512*2 | 0.51 |
| SO-Net [42] | 512*2 | 0.44 |
| SCN [39] | 256*2 | 0.60 |

## 4.3 Object segmentation evaluation On ShapeNet Parts

Objects represented with 3D point clouds have made the segmentation more complicated because of confusing point clouds to noisy points. Therefore, the point cloud preprocessing step should precede the segmentation process. The proposed object segmentation methods on 3D point clouds were trained and evaluated on ShapeNet parts. Table 7 compares all of these methods with 16 object categories and a presentation of the best results for each category.

In Table 7, we find that some methods reached a good performance for the 16 categories like RS-CNN, LSDTM-GCNN, VoxRec, and DC-GNN, which hit the best results. While other methods only achieved high accuracies for selected categories. In terms of Mean values we find that SFCNN [55] came in second place by a value of 85.4% with a difference of 0.8% as opposed to RS-CNN [50]. From the means values we notice that all results are close and the difference between the highest and lowest values is only 5.3%.

Table 7: Object part segmentation results on ShapeNetPart dataset.

| Method | Mean | Aero | Bag | Cap | Chair | Car | Ear-ph | Gui | Knife | Lamp | Laptop | Motor | Mug | Pistol | Rocket | Skate | Table |
|---|---|---|---|---|---|---|---|---|---|---|---|---|---|---|---|---|---|

| Method | | | | | | | | | | | | | | | | |
|---|---|---|---|---|---|---|---|---|---|---|---|---|---|---|---|---|
| PointNet [33] | 83.7 | 83.4 | 78.7 | 82.5 | 74.9 | 89.6 | 73.0 | 91.5 | 85.9 | 80.8 | 95.3 | 65.2 | 93.0 | 81.2 | 57.9 | 72.8 | 80.6 |
| VoxRec [61] | 81.5 | 75.6 | 83.0 | 83.4 | 83.9 | 73.9 | 75.3 | 88.6 | 88.4 | 73.3 | 94.5 | 72.6 | 96.5 | 80.3 | 69.1 | 83.8 | - |
| SO-Net [42] | 84.6 | 81.9 | 83.5 | 84.8 | 78.1 | 90.8 | 72.2 | 90.1 | 83.6 | 82.3 | 95.2 | 69.3 | 94.2 | 80.0 | 51.6 | 72.1 | 82.6 |
| RS-CNN [50] | 86.2 | 83.5 | 84.8 | 88.8 | 79.6 | 91.2 | 81.1 | 91.6 | 88.4 | 86.0 | 96.0 | 73.7 | 94.1 | 83.4 | 60.5 | 77.7 | 83.6 |
| SFCNN [55] | 85.4 | 83.0 | 83.4 | 87.0 | 80.2 | 90.1 | 75.9 | 91.1 | 86.2 | 84.2 | 96.7 | 69.5 | 94.8 | 82.5 | 59.9 | 75.1 | 82.9 |
| A-SCN [39] | 84.6 | 83.8 | 80.8 | 83.5 | 79.3 | 90.5 | 69.8 | 91.7 | 86.5 | 82.9 | 96.0 | 69.2 | 93.8 | 82.5 | 62.9 | 74.4 | 80.8 |
| DGCNN [58] | 85.2 | 84.0 | 83.4 | 86.7 | 77.8 | 90.6 | 74.7 | 91.2 | 87.5 | 82.8 | 95.7 | 66.3 | 94.9 | 81.1 | 63.5 | 74.5 | 82.6 |
| LSDTM-GCNN [97] | - | 84.4 | 86.6 | 90.2 | 79.9 | 91.9 | 80.6 | 95.3 | 90.0 | 83.2 | 96.1 | 71.7 | 95.2 | 80.3 | 68.1 | 79.3 | 97.2 |
| DC-GNN [71] | - | 85.3 | 82.2 | 86.0 | 81.0 | 91.3 | 78.9 | 92.2 | 88.0 | 85.0 | 95.8 | 72.8 | 95.2 | 84.5 | 54.2 | 76.9 | 82.9 |
| AGNet [98] | - | 84.1 | 83.2 | 86.0 | 78.8 | 90.6 | 76.9 | 91.9 | 88.4 | 82.3 | 96.0 | 65.5 | 93.7 | 84.2 | 64.2 | 76.8 | 80.6 |
| DANet [96] | - | 83.9 | 83.2 | 85.0 | 79.7 | 91.1 | 77.3 | 91.9 | 88.4 | 84.8 | 95.7 | 71.9 | 94.7 | 83.2 | 58.0 | 75.1 | 82.8 |
| 3D-SelfCutMix [100] | - | 82.4 | 72.0 | 79.1 | 76.9 | 90.6 | 75.2 | 91.6 | 88.6 | 85.3 | 96.1 | 67.8 | 92.0 | 79.2 | 57.8 | 72.9 | 83.5 |

Also for some categories like Motor, Rocket, and Skate, results are lower than some other categories like Car, Gui, Mug, and Knife. Figure 6 illustrates some examples of the segmented part of objects from the ShapNet Part dataset obtained using the VoxRec method.

## 4.4 3D object reconstruction evaluation

To evaluate 3D object reconstruction methods, authors used ShapeNet dataset which contains segmented features. The metrics used for this evaluation for 3D point cloud reconstruction are Chamfer distance (CD), and Earth mover's distance (EMD). These metrics are defined in [77] as follows:

$$D(P, \hat{P}) = \frac{1}{|P|} \sum_{p \in P} \min_{\hat{p} \in \hat{P}} \|p - \hat{p}\|_2^2 + \frac{1}{|\hat{P}|} \sum_{\hat{p} \in \hat{P}} \min_{p \in P} \|p - \hat{p}\|_2^2$$

While given predicted point clouds $\hat{p} \in \hat{P}$ and the ground truth $p \in P$.

$$EMD(P, \hat{P}) = \min_{\phi: P \to \hat{P}} \frac{1}{|P|} \sum_{p \in P} \|p - \phi p\|$$

Where $\phi: P \to \hat{P}$ denotes a bijection that minimizes the average distance between corresponding points.

3D object reconstruction methods were trained and tested on ShapeNet dataset. This dataset includes synthetic images from different viewpoints, wherein 13 categories are considered during the evaluation process. Individual results are presented in Table 8 fr each category. From the Table, we find that some methods generated point clouds

Table 8: Quantitative comparison of different object reconstruction proposed models on ShapeNet.

| Category | CD | | | | EMD | | | |
|---|---|---|---|---|---|---|---|---|
| | [76] | [75] | [77] | [78] | [76] | [75] | [77] | [78] |
| plane | 0.430 | 0.477 | 0.386 | 0.329 | 0.396 | 0.579 | 0.527 | 0.382 |
| bench | 0.629 | 0.624 | 0.436 | 0.459 | 1.113 | 0.965 | 0.815 | 0.431 |
| cabinet | 0.439 | 0.381 | 0.373 | 0.607 | 2.986 | 2.563 | 2.147 | 0.494 |
| car | 0.333 | 0.268 | 0.308 | 0.439 | 1.747 | 1.297 | 1.306 | 0.361 |
| chair | 0.645 | 0.610 | 0.606 | 0.48 | 1.946 | 1.399 | 1.257 | 0.645 |
| monitor | 0.722 | 0.755 | 0.501 | 0.58 | 1.891 | 1.536 | 1.314 | 0.845 |
| lamp | 1.193 | 1.295 | 0.969 | 0.639 | 1.222 | 1.314 | 1.007 | 0.594 |
| speaker | 0.756 | 0.739 | 0.632 | 2.89 | 3.490 | 2.951 | 2.441 | 0.425 |
| firearm | 0.423 | 0.453 | 0.463 | 0.585 | 0.397 | 0.667 | 0.572 | 0.503 |
| couch | 0.549 | 0.490 | 0.439 | 0.390 | 2.207 | 1.642 | 1.536 | 0.737 |
| table | 0.517 | 0.498 | 0.589 | 0.626 | 2.121 | 1.480 | 1.340 | 0.605 |
| cellphone | 0.438 | 0.421 | 0.332 | 0.427 | 1.019 | 0.724 | 0.674 | 0.377 |
| watercraft | 0.633 | 0.670 | 0.478 | 0.555 | 0.945 | 0.814 | 0.730 | 0.489 |
| mean | 0.593 | 0.591 | 0.501 | 0.538 | 1.653 | 1.380 | 1.205 | 0.530 |

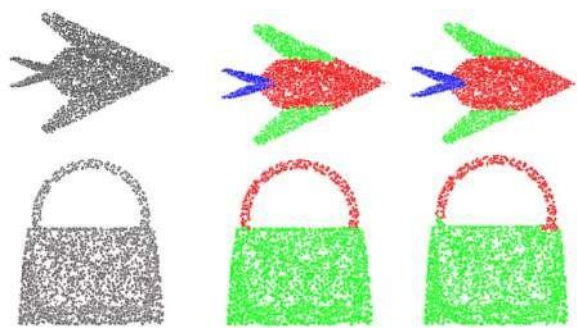

Figure 6: Segmented results of some example from Shapenet Parts dataset using the method in [61] (VoxRec). Left: 3D point cloud. Middle: Ground-truth. Right: the segmented results

for 3D objects with considerable accuracy using the two metrics CD and EMD. This translates into generating completed shapes and surfaces despite the presence of varied objects in the dataset. The comparison of relevant results reported in Table 8, shows that for CD metric the [76] method reached the highest values for all categories. While [75] method reached the second-best results for some categories like Plan, Monitor, Lamp, and Watercraft. Also, [78] yielded the best results for many categories like Cabinet, Car, Speaker, Firearm and Table ( see Figure 7). When it comes to EMD metric, [76] reached the highest results for all categories except for Lamp, Speaker, and Firearm. While [76] method came in the second place for most categories. [78] results using EMD are the lowest compared to previous methods with a difference of 0.01 to 2.5 points. To summarize, it is clear that the experimental results are convincing for certain categories yet need more work for others.

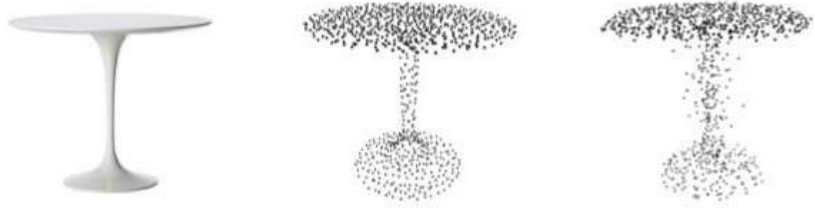

Figure 7: Generated 3D point cloud table using the method in [78]. Left: original image. Middle: Ground-truth. Right: the generated table

-0cm

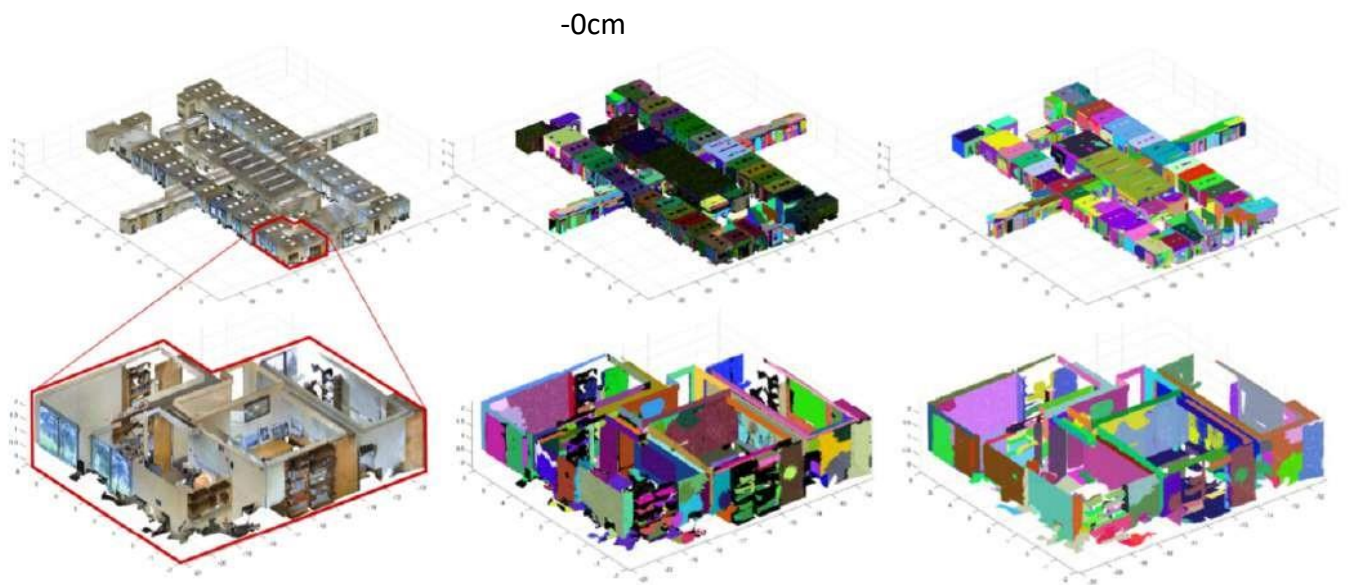

Figure 8: The obtained results of [106] on 2D-3D-S dataset

As an illustration of 3D object reconstruction, figure 8 represents the output of [106] approach applied to 2D-3D-Semantics Dataset. The evaluation was carried out on a set composed of a tile size of 1 million 3D points runnable on a normal machine CPU (the tile size can be bigger if working with a high machine performance). The figure reveals that some parts are not well segmented due to the noise and occlusions in the dataset. Besides, due to feature variance, some parts are over-segmented.

# 5   Challenges and future directions

Point cloud representation is a new type of data that has been used to support and enhance many computer-aided applications. It comes with various benefits in terms of data transfer cost, efficient shredding data including multi-view data representation of objects, and the ability to reconstruct and generate real objects from point cloud data. However, this technique often faces challenges pertaining essentially to data analysis due to the presence of noisy points, blind and cluttered scenes, uncompleted parts of the objects, etc. In the following, we will present current issues of the 3D point cloud as well as some of the solutions and future trends intended to solve each one of them.

**Incomplete, noisy, and unstructured data:**

Using dots of XYZ coordinates can represent a 3D synthetic object , but using a sensor like LiDAR, these coordinates cannot be perfectly represented due to the number of point (dots) captured as well as the distribution of these points. The number of point generated can be considered as a cloud of noise while there are many objects in complex scene. For the same reason, the gathered data can contain some uncompleted part of the objects. These parts can be important for recognizing the object in order to segment it or reconstruct it. Also using LiDAR sensor, the generated point cloud represent just one side of the object. Also, we can find that some regions have sparse swarm of dots while others have dense dots which make the data irregular because the distance between the points is varying due to fact that each point is scanned independently.

In order to overcome this problem, researchers attempted to use deep learning techniques to complete the missing parts of 3D point cloud objects with reconstruction methods [107]. Also, the proposed methods tried to recognize the objects even if represented with a minimum swarm of dots.

Another issue facing deep learning methods (specifically the ones including convolutional neural networks (CNN)) is that they are usually trained on structured, ordered and regular data. So, to face 3D point cloud data discrepancy, researchers decided to convert the point cloud into a structured grid presentation to serve as input for their models.

**Hard to store and process:**

LiDAR data that contains the information of the 3D objects (represented with points in 3D space), can result into massive files to be stored and transferred [108]. This tremendous amount of information makes the data hard to process and to handle. When it comes to training a model with a large-scale dataset, the process is costly in terms of memory and demands high performance machines. Nevertheless, the use of multi-dimensional vectors or tensors of the point cloud reduce the size of each file and enables the machine process them more easily [109]. Also, the use of compact representations by encoding 3D shapes can make the classification and recognition tasks more doable [110].

# 6  Conclusion

3D point cloud is a new representation of data beside the existing ones such as image, video, and time-series data. This presentation can give many information using swarm dots about the object or the scenes represented in 3D shapes. Working on this data for recognizing the object segmentation the part of these object or also reconstruct objects and scenes is a new breakthrough in computer vision. It has opened multiple opportunities in different research and development fields, wherein the complete information about the objects is required, such as autonomous driving, remote sensing images mapping, and object and scenes reconstruction, etc. To inform the state-of-the-art, we proposed, an extensive critical review of the tasks performed on 3D point cloud, which has been designed following a well defined methodology. Accordingly, different tasks was firstly presented. Next, we listed the datasets used for performance evaluation. Also, we drew a comparison of the obtained results for each task using different evaluation metrics. A list of challenges of the methods working on 3D point cloud were provided with potential solutions and futures directions.

## Acknowledgments

This research work was made possible by research grant support (QUEX-CENG-SCDL-19/20-1 ) from Supreme Committee for Delivery and Legacy (SC) in Qatar.

## References

[1] H. Saffi, Y. Hmamouche, O. Elharrouss, A. E. F. Seghrouchni, Inception-based deep learning architecture for 3d point cloud completion, in: 2022 18th IEEE International Conference on Advanced Video and Signal Based Surveillance (AVSS), IEEE, 2022, pp. 1–7.

[2] R. Rusu, S. Cousins, 3d is here: Point cloud library (pcl), 2011.  doi:10.1109/ICRA.2011.5980567.

[3] D. Gao, J. Liu, R. Wu, D. Cheng, X. Fan, X. Tang, Utilizing relevant rgb–d data to help recognize rgb images in the target domain, International Jour-

nal of Applied Mathematics and Computer Science 29 (2019) 611–621. doi: 10.2478/amcs-2019-0045.

[4] G. Ping, H. Wang, 3d reconstruction from a single image (2019) 47–52 doi:10.1109/CIS-RAM47153.2019.9095839.

[5] V. Narayanan, M. Likhachev, Perch: Perception via search for multi-object recognition and localization (10 2015).

[6] D. Heisterkamp, P. Bhattacharya, Matching of 3d polygonal arcs., IEEE Trans. Pattern Anal. Mach. Intell. 19 (1997) 68–73. doi:10.1109/34.566813.

[7] S. Li, On topological consistency and realization, Constraints 11 (2006) 31–51. doi:10.1007/s10601-006-6847-9.

[8] S. Daftry, C. Hoppe, H. Bischof, Building with drones: Accurate 3d facade reconstruction using mavs, in: 2015 IEEE International Conference on Robotics and Automation (ICRA), 2015, pp. 3487–3494.

[9] J. Han, L. Shao, D. Xu, J. Shotton, Enhanced computer vision with microsoft kinect sensor: A review, IEEE Transactions on Cybernetics 43 (5) (2013) 1318–1334.

[10] W. Xu, I.-S. Lee, S.-H. Lee, B. Lu, E.-J. Lee, Multiview-based hand posture recognition method based on point cloud, KSII Transactions on Internet and Information Systems 9 (2015) 2585–2598. doi:10.3837/tiis.2015.07.0014.

[11] J. Park, H. Kim, Yu-Wing Tai, M. S. Brown, I. Kweon, High quality depth map upsampling for 3d-tof cameras, in: 2011 International Conference on Computer Vision, 2011, pp. 1623–1630.

[12] N. B. Zaman F., Wong Y.P., Multiview-based hand posture recognition method based on point cloud, Springer, Singapore 398 (09 2017). doi:https://doi.org/10.1007/978-981-10-1721-6_31.

[13] Y. Tian, W. Song, S. Sun, S. Fong, S. Zou, 3d object recognition method with multiple feature extraction from lidar point clouds, The Journal of Supercomputing (2019) 1–13.

[14] W. Wang, S. Zhou, J. Li, X. Li, J. Yuan, Z. Jin, Temporal pulses driven spiking neural network for fast object recognition in autonomous driving (01 2020).

[15] H. Kasaei, M. Ghorbani, J. Schilperoort, W. Rest, Investigating the importance of shape features, color constancy, color spaces and similarity measures in open-ended 3d object recognition (02 2020).

[16] S. Dong, L. Ren, K. Zou, W. Li, 3d object recognition method based on point cloud sequential coding, 2020, pp. 297–300. doi:10.1145/3383972.3383984.

[17] N. Akai, T. Hirayama, H. Murase, Semantic localization considering uncertainty of object recognition (05 2020). doi:10.1109/LRA.2020.2998403.

[18] O. Elharrouss, Y. Hmamouche, A. K. Idrissi, B. El Khamlichi, A. El Fallah-Seghrouchni, Refined edge detection with cascaded and high-resolution convolutional network, Pattern Recognition 138 (2023) 109361.

[19] M. Elasri, O. Elharrouss, S. Al-Maadeed, H. Tairi, Image generation: A review, Neural Processing Letters 54 (5) (2022) 4609–4646.

[20] O. Elharrouss, N. Almaadeed, S. Al-Maadeed, F. Khelifi, Pose-invariant face recognition with multitask cascade networks, Neural Computing and Applications (2022) 1–14.

[21] O. Elharrouss, Y. Akbari, N. Almaadeed, S. Al-Maadeed, Backbones-review: Feature extraction networks for deep learning and deep reinforcement learning approaches, arXiv preprint arXiv:2206.08016 (2022).

[22] O. Elharrouss, N. Almaadeed, S. Al-Maadeed, A review of video surveillance systems, Journal of Visual Communication and Image Representation (2021) 103116.

[23] Y. Akbari, N. Almaadeed, S. Al-maadeed, O. Elharrouss, Applications, databases and open computer vision research from drone videos and images: a survey, Artificial Intelligence Review 54 (5) (2021) 3887–3938.

[24] U. Asif, M. Bennamoun, F. A. Sohel, Rgb-d object recognition and grasp detection using hierarchical cascaded forests, IEEE Transactions on Robotics 33 (3) (2017) 547–564.

[25] A. G. Buch, L. Kiforenko, D. Kraft, Rotational subgroup voting and pose clustering for robust 3d object recognition, in: 2017 IEEE International Conference on Computer Vision (ICCV), 2017, pp. 4137–4145.

[26] A. Borcs, B. Nagy, C. Benedek, Instant object detection in lidar point clouds, IEEE Geoscience and Remote Sensing Letters 14 (7) (2017) 992–996.

[27] P. Falco, S. Lu, A. Cirillo, C. Natale, S. Pirozzi, D. Lee, Cross-modal visuo-tactile object recognition using robotic active exploration, in: 2017 IEEE International Conference on Robotics and Automation (ICRA), 2017, pp. 5273–5280.

[28] D. Fehr, W. Beksi, D. Zermas, N. Papanikolopoulos, Covariance based point cloud descriptors for object detection and recognition, Computer Vision and Image Understanding 142 (01 2016). doi:10.1016/j.cviu.2015.06.008.

[29] F. Gomez-Donoso, A. Garcia-Garcia, J. Garcia-Rodriguez, S. Orts-Escolano, M. Cazorla, Lonchanet: A sliced-based cnn architecture for real-time 3d object recognition, in: 2017 International Joint Conference on Neural Networks (IJCNN), 2017, pp. 412–418.

[30] S. H. Kasaei, A. M. Tomé, L. Seabra Lopes, M. Oliveira, Good: A global orthographic object descriptor for 3d object recognition and manipulation, Pattern Recognition Letters 83 (2016) 312 – 320, efficient Shape Representation, Matching, Ranking, and its Applications. doi:https://doi.org/10.1016/j.patrec.2016.07.006.
URL http://www.sciencedirect.com/science/article/pii/S0167865516301684

[31] S. Luo, W. Mou, K. Althoefer, H. Liu, Iterative closest labeled point for tactile object shape recognition, in: 2016 IEEE/RSJ International Conference on Intelligent Robots and Systems (IROS), 2016, pp. 3137–3142.

[32] B. Li, 3d fully convolutional network for vehicle detection in point cloud, 2017, pp. 1513–1518. doi:10.1109/IROS.2017.8205955.

[33] C. R. Qi, H. Su, K. Mo, L. J. Guibas, Pointnet: Deep learning on point sets

for 3d classification and segmentation, in: Proceedings of the IEEE conference on computer vision and pattern recognition, 2017, pp. 652–660.

[34] C. R. Qi, L. Yi, H. Su, L. J. Guibas, Pointnet++: Deep hierarchical feature learning on point sets in a metric space, arXiv preprint arXiv:1706.02413 (2017).

[35] Z. Gojcic, C. Zhou, J. D. Wegner, A. Wieser, The perfect match: 3d point cloud matching with smoothed densities, in: Proceedings of the IEEE/CVF Conference on Computer Vision and Pattern Recognition, 2019, pp. 5545–5554.

[36] J. Wang, R. Lindenbergh, M. Menenti, Sigvox – a 3d feature matching algorithm for automatic street object recognition in mobile laser scanning point clouds, ISPRS Journal of Photogrammetry and Remote Sensing 128 (2017) 111–129. doi:10.1016/j.isprsjprs.2017.03.012.

[37] J. Rangel Ortiz, J. Martínez-Gómez, C. Gonzalez, I. García-Varea, M. Cazorla, Semi-supervised 3d object recognition through cnn labeling, Applied Soft Computing 65 (2018) 603–613. doi:10.1016/j.asoc.2018.02.005.

[38] M. A. Uy, G. H. Lee, Pointnetvlad: Deep point cloud based retrieval for large-scale place recognition (2018). arXiv:1804.03492.

[39] S. Xie, S. Liu, Z. Chen, Z. Tu, Attentional shapecontextnet for point cloud recognition, 2018, pp. 4606–4615. doi:10.1109/CVPR.2018.00484.

[40] H. You, Y. Feng, R. Ji, Y. Gao, Pvnet: A joint convolutional network of point cloud and multi-view for 3d shape recognition (08 2018).

[41] Y. Zhou, O. Tuzel, Voxelnet: End-to-end learning for point cloud based 3d object detection (2018) 4490–4499.

[42] J. Li, B. Chen, G. Lee, So-net: Self-organizing network for point cloud analysis, 2018. doi:10.1109/CVPR.2018.00979.

[43] A. Ahmed, A. Jalal, K. Kim, Rgb-d depth images for object segmentation, localization and recognition in indoor scenes using feature descriptor and hough voting, 2019. doi:10.1109/IBCAST47879.2020.9044545.

[44] J. Chen, Z. Kira, Y. Cho, Deep learning approach to point cloud scene understanding for automated scan to 3d reconstruction, Journal of Computing in Civil Engineering 33 (05 2019). doi:10.1061/(ASCE)CP.1943-5487.0000842.

[45] A. Cheraghian, S. Rahman, L. Petersson, Zero-shot learning of 3d point cloud objects, in: 2019 16th International Conference on Machine Vision Applications (MVA), 2019, pp. 1–6.

[46] A. Cheraghian, S. Rahman, D. Campbell, L. Petersson, Transductive zero-shot learning for 3d point cloud classification, in: Proceedings of the IEEE/CVF Winter Conference on Applications of Computer Vision (WACV), 2020.

[47] Q. Huang, Y. Wang, Z. Yin, View-based weight network for 3d object recognition, Image and Vision Computing 93 (2020) 103828. doi:https://doi.org/10.1016/j.imavis.2019.11.006.
URL http://www.sciencedirect.com/science/article/pii/S0262885619304214

[48] R. Li, X. Li, C.-W. Fu, D. Cohen-Or, P.-A. Heng, Pu-gan: A point cloud upsampling adversarial network, 2019, pp. 7202–7211. doi:10.1109/ICCV.2019.00730.

[49] H. Liu, Y. Cong, C. Yang, Y. Tang, Efficient 3d object recognition via geometric information preservation, Pattern Recognition 92 (03 2019). doi:10.1016/j.patcog.2019.03.025.

[50] Y. Liu, B. Fan, S. Xiang, C. Pan, Relation-shape convolutional neural network for point cloud analysis, in: 2019 IEEE/CVF Conference on Computer Vision and Pattern Recognition (CVPR), 2019, pp. 8887–8896.

[51] X. Liu, M. Yan, J. Bohg, Meteornet: Deep learning on dynamic 3d point cloud sequences, in: Proceedings of the IEEE/CVF International Conference on Computer Vision (ICCV), 2019.

[52] Z. Luo, J. Li, Z. Xiao, Z. G. Mou, X. Cai, C. Wang, Learning high-level features by fusing multi-view representation of mls point clouds for 3d object recognition in road environments, ISPRS Journal of Photogrammetry and Remote Sensing 150 (2019) 44 – 58. doi:https://doi.org/10.1016/j.isprsjprs.2019.01.024.

URL http://www.sciencedirect.com/science/article/pii/ S09242716193 00322

[53] J. Mao, X. Wang, H. Li, Interpolated convolutional networks for 3d point cloud understanding, 2019, pp. 1578–1587. doi:10.1109/ICCV.2019.00166.

[54] A. Paigwar, O. Erkent, C. Wolf, C. Laugier, Attentional pointnet for 3d-object detection in point clouds, in: 2019 IEEE/CVF Conference on Computer Vision and Pattern Recognition Workshops (CVPRW), 2019, pp. 1297–1306.

[55] Y. Rao, J. Lu, J. Zhou, Spherical fractal convolutional neural networks for point cloud recognition, in: 2019 IEEE/CVF Conference on Computer Vision and Pattern Recognition (CVPR), 2019, pp. 452–460.

[56] H. Wang, X. Lou, Y. Cai, Y. Li, L. Chen, Real-time vehicle detection algorithm based on vision and lidar point cloud fusion, Journal of Sensors 2019 (2019) 1–9. doi:10.1155/2019/8473980.

[57] Z. Yang, Y. Sun, S. Liu, X. Shen, J. Jia, Std: Sparse-to-dense 3d object detector for point cloud, in: 2019 IEEE/CVF International Conference on Computer Vision (ICCV), 2019, pp. 1951–1960.

[58] Y. Wang, Y. Sun, Z. Liu, S. E. Sarma, M. M. Bronstein, J. M. Solomon, Dynamic graph cnn for learning on point clouds, Acm Transactions On Graphics (tog) 38 (5) (2019) 1–12.

[59] Y. Zhou, P. Sun, Y. Zhang, D. Anguelov, J. Gao, T. Ouyang, J. Guo, J. Ngiam, V. Vasudevan, End-to-end multi-view fusion for 3d object detection in lidar point clouds (10 2019).

[60] D. Li, H. Wang, N. Liu, X. Wang, J. Xu, 3d object recognition and pose estimation from point cloud using stably observed point pair feature, IEEE Access 8 (2020) 44335–44345.

[61] A. Karambakhsh, B. Sheng, P. Li, P. Yang, Y. Jung, D. D. Feng, Voxrec: Hybrid convolutional neural network for active 3d object recognition, IEEE Access 8 (2020) 70969–70980.

[62] Q. Meng, W. Wang, T. Zhou, J. Shen, D. Dai, Weakly supervised 3d object detection from lidar point cloud (07 2020).

[63] N. Sedaghat, M. Zolfaghari, T. Brox, Orientation-boosted voxel nets for 3d object recognition (04 2016).

[64] G. Melotti, C. Premebida, N. Gonçalves, Multimodal deep-learning for object recognition combining camera and lidar data, in: 2020 IEEE International Conference on Autonomous Robot Systems and Competitions (ICARSC), 2020, pp. 177–182.

[65] Y. Yang, F. Chen, F. Wu, D. Zeng, Y.-m. Ji, X.-Y. Jing, Multi-view semantic learning network for point cloud based 3d object detection, Neurocomputing 397 (03 2020). doi:10.1016/j.neucom.2019.10.116.

[66] K. Zhang, S. Wang, L. Ji, C. Wang, Dnn based camera and lidar fusion framework for 3d object recognition, Journal of Physics: Conference Series 1518 (2020) 012044. doi:10.1088/1742-6596/1518/1/012044.

[67] J. Huang, S. You, Point cloud labeling using 3d convolutional neural network, 2016. doi:10.1109/ICPR.2016.7900038.

[68] Z. Lakhili, A. El Alami, A. Mesbah, A. Berrahou, H. Qjidaa, Rigid and non-rigid 3d shape classification based on 3d hahn moments neural networks model, Multimedia Tools and Applications (2022) 1–24.

[69] Y. Tian, W. Song, L. Chen, S. Fong, Y. Sung, J. Kwak, A 3d object recognition method from lidar point cloud based on usae-bls, IEEE Transactions on Intelligent Transportation Systems (2022).

[70] A. S. Gezawa, Z. A. Bello, Q. Wang, L. Yunqi, A voxelized point clouds representation for object classification and segmentation on 3d data, The Journal of Supercomputing 78 (1) (2022) 1479–1500.

[71] M. Meraz, M. A. Ansari, M. Javed, P. Chakraborty, Dc-gnn: drop channel graph neural network for object classification and part segmentation in the point cloud, International Journal of Multimedia Information Retrieval 11 (2) (2022) 123–133.

[72] S. Chen, T. Yu, P. Li, R2-mlp: Round-roll mlp for multi-view 3d object recognition, arXiv preprint arXiv:2211.11085 (2022).

[73] R. Xu, W. Ma, Q. Mil, H. Zha, Reinview: Re-interpreting views for multi-view 3d object recognition, in: 2022 IEEE/RSJ International Conference on Intelligent Robots and Systems (IROS), IEEE, 2022, pp. 6630–6636.

[74] Y. Lu, F. Yang, F. Chen, D. Xie, Pic-net: Point cloud and image collaboration network for large-scale place recognition (08 2020).

[75] N. Wang, Y. Zhang, Z. Li, Y. Fu, W. Liu, Y.-G. Jiang, Pixel2mesh: Generating 3d mesh models from single rgb images, in: Proceedings of the European Conference on Computer Vision (ECCV), 2018, pp. 52–67.

[76] H. Fan, H. Su, L. J. Guibas, A point set generation network for 3d object reconstruction from a single image, in: Proceedings of the IEEE conference on computer vision and pattern recognition, 2017, pp. 605–613.

[77] C. Zou, D. Hoiem, Silhouette guided point cloud reconstruction beyond occlusion, in: Proceedings of the IEEE/CVF Winter Conference on Applications of Computer Vision, 2020, pp. 41–50.

[78] A. J. Afifi, J. Magnusson, T. A. Soomro, O. Hellwich, Pixel2point: 3d object reconstruction from a single image using cnn and initial sphere, IEEE Access 9 (2020) 110–121.

[79] P. Jin, S. Liu, J. Liu, H. Huang, L. Yang, M. Weinmann, R. Klein, Weakly-supervised single-view dense 3d point cloud reconstruction via differentiable renderer, Chinese Journal of Mechanical Engineering 34 (1) (2021) 1–11.

[80] Y. Nie, J. Hou, X. Han, M. Nießner, Rfd-net: Point scene understanding by semantic instance reconstruction, in: Proceedings of the IEEE/CVF Conference on Computer Vision and Pattern Recognition, 2021, pp. 4608–4618.

[81] A. Dai, A. X. Chang, M. Savva, M. Halber, T. Funkhouser, M. Nießner, Scannet: Richly-annotated 3d reconstructions of indoor scenes, in: Proceedings of the IEEE conference on computer vision and pattern recognition, 2017, pp. 5828–5839.

[82] A. Avetisyan, M. Dahnert, A. Dai, M. Savva, A. X. Chang, M. Nießner, Scan2cad: Learning cad model alignment in rgb-d scans, in: Proceedings of the IEEE/CVF Conference on Computer Vision and Pattern Recognition, 2019, pp. 2614–2623.

[83] M. Bassier, M. Vergauwen, Topology reconstruction of bim wall objects from point cloud data, Remote Sensing 12 (11) (2020) 1800.

[84] M. Bassier, M. Vergauwen, Unsupervised reconstruction of building information modeling wall objects from point cloud data, Automation in Construction 120 (2020) 103338.

[85] J. Han, M. Rong, H. Jiang, H. Liu, S. Shen, Vectorized indoor surface reconstruction from 3d point cloud with multistep 2d optimization, ISPRS Journal of Photogrammetry and Remote Sensing 177 (2021) 57–74.

[86] J. Chen, Z. Kira, Y. K. Cho, Deep learning approach to point cloud scene understanding for automated scan to 3d reconstruction, Journal of Computing in Civil Engineering 33 (4) (2019) 04019027.

[87] R. Cupec, I. Vidović, D. Filko, P. urović, Object recognition based on convex hull alignment, Pattern Recognition 102 (2020) 107199. doi:10.1016/j.patcog.2020.107199.

[88] H. Zhao, M. Tang, H. Ding, Hoppf: A novel local surface descriptor for 3d object recognition, Pattern Recognition 103 (2020) 107272. doi:https://doi.org/10.1016/j.patcog.2020.107272.
URL http://www.sciencedirect.com/science/article/pii/S0031320320300777

[89] J. Yang, L. Song, S. Liu, Z. Li, X. Li, H. Sun, J. Sun, N. Zheng, Dbq-ssd: Dynamic ball query for efficient 3d object detection, arXiv preprint arXiv:2207.10909 (2022).

[90] Q. Xu, Y. Zhong, U. Neumann, Behind the curtain: Learning occluded shapes for 3d object detection, in: Proceedings of the AAAI Conference on Artificial Intelligence, Vol. 36, 2022, pp. 2893–2901.

[91] Z. Li, Y. Yao, Z. Quan, J. Xie, W. Yang, Spatial information enhancement network for 3d object detection from point cloud, Pattern Recognition 128 (2022) 108684.

[92] R. Qian, X. Lai, X. Li, Badet: Boundary-aware 3d object detection from point clouds, Pattern Recognition 125 (2022) 108524.

[93] Z. Liu, T. Huang, B. Li, X. Chen, X. Wang, X. Bai, Epnet++: Cascade bi-directional fusion for multi-modal 3d object detection, IEEE Transactions on Pattern Analysis and Machine Intelligence (2022).

[94] J. Li, R. Li, J. Li, J. Wang, Q. Wu, X. Liu, Dual-view 3d object recognition and detection via lidar point cloud and camera image, Robotics and Autonomous Systems 150 (2022) 103999.

[95] O. Elharrouss, S. Al-Maadeed, N. Subramanian, N. Ottakath, N. Almaadeed, Y. Himeur, Panoptic segmentation: A review, arXiv preprint arXiv:2111.10250 (2021).

[96] C. Zhou, Y. Xie, X. He, T. Yuan, Q. Ling, Dual attention network for point cloud classification and segmentation, in: 2022 41st Chinese Control Conference (CCC), IEEE, 2022, pp. 6482–6486.

[97] W. J. Zhang, S. Z. Su, Q. Q. Hong, B. Z. Wang, L. Sun, Long short-distance topology modelling of 3d point cloud segmentation with a graph convolution neural network, IET Computer Vision (2022).

[98] W. Jing, W. Zhang, L. Li, D. Di, G. Chen, J. Wang, Agnet: An attention-based graph network for point cloud classification and segmentation, Remote Sensing 14 (4) (2022) 1036.

[99] M. Sheikh, M. A. Asghar, R. Bibi, M. N. Malik, M. Shorfuzzaman, R. M. Mehmood, S.-H. Kim, Dft-net: Deep feature transformation based network for object categorization and part segmentation in 3-dimensional point clouds, Sensors 22 (7) (2022) 2512.

[100] Y.-Y. Xu, Y.-Y. Ji, S.-Y. Huang, Z.-H. Lin, Y.-C. F. Wang, 3d-selfcutmix: Self-supervised learning for 3d point cloud analysis, in: 2022 IEEE International Conference on Image Processing (ICIP), IEEE, 2022, pp. 676–680.

[101] Q. Wang, Y. Tan, Z. Mei, Computational methods of acquisition and processing of 3d point cloud data for construction applications, Archives of Computational Methods in Engineering 27 (2) (2020) 479–499. doi:10.1007/s11831-019-09320-4. URL https://app.dimensions.ai/details/publication/pub.1112262162

[102] J. Deng, W. Dong, R. Socher, L.-J. Li, K. Li, L. Fei-Fei, Imagenet: A large-scale hierarchical image database, in: 2009 IEEE conference on computer vision and pattern recognition, Ieee, 2009, pp. 248–255.

[103] Q. Cao, L. Shen, W. Xie, O. M. Parkhi, A. Zisserman, Vggface2: A dataset for recognising faces across pose and age, in: 2018 13th IEEE international conference on automatic face & gesture recognition (FG 2018), IEEE, 2018, pp. 67–74.

[104] L. Wang, Z. Song, X. Zhang, C. Wang, G. Zhang, L. Zhu, J. Li, H. Liu, Sat-gcn: Self-attention graph convolutional network-based 3d object detection for autonomous driving, Knowledge-Based Systems 259 (2023) 110080.

[105] W. Wang, G. Chen, H. Zhou, X. Wang, Ovpt: Optimal viewset pooling transformer for 3d object recognition, in: Proceedings of the Asian Conference on Computer Vision, 2022, pp. 4444–4461.

[106] M. Bassier, M. Bonduel, B. Van Genechten, M. Vergauwen, Segmentation of large unstructured point clouds using octree-based region growing and conditional random fields, The International Archives of the Photogrammetry, Remote Sensing and Spatial Information Sciences 42 (2W8) (2017) 25–30.

[107] P. Spurek, A. Kasymov, M. Mazur, D. Janik, S. Tadeja, Ł. Struski, J. Tabor, T. Trzciński, Hyperpocket: Generative point cloud completion, arXiv preprint arXiv:2102.05973 (2021).

[108] P. Spurek, M. Zieba, J. Tabor, T. Trzciński, Hyperflow: Representing 3d objects as surfaces, arXiv preprint arXiv:2006.08710 (2020).

[109] M. Zamorski, M. Zieba, P. Klukowski, R. Nowak, K. Kurach, W. Stokowiec, T. Trzciński, Adversarial autoencoders for compact representations of 3d point clouds, Computer Vision and Image Understanding 193 (2020) 102921.

[110] K. Kania, M. Zieba, T. Kajdanowicz, Ucsg-net–unsupervised discovering of constructive solid geometry tree, arXiv preprint arXiv:2006.09102 (2020).